%% file: main.tex
\documentclass[journal]{IEEEtran}
\usepackage{cite}
\usepackage{amsmath,amssymb,amsfonts, amsthm}
 \usepackage{algorithmic}
\usepackage{algorithm}
\usepackage{bbm}
\usepackage{graphicx}
\usepackage{textcomp}
\usepackage{xcolor}
\usepackage{kotex}
\usepackage{subfigure}
\usepackage{booktabs}
\usepackage{multicol}
\usepackage{multirow}
\usepackage[normalem]{ulem}
\usepackage{lettrine}

\usepackage{xspace}
\usepackage{amsmath}
\usepackage{amsfonts}

\newcommand{\BfPara}[1]{{\noindent\bf#1.}\xspace}

\begin{document}

\title{Realizing Stabilized Landing for Computation-Limited Reusable Rockets: A Quantum Reinforcement Learning Approach}

\author{
    Gyu Seon Kim, 
    JaeHyun Chung, and
    Soohyun Park

    \thanks{G. S. Kim, J. Chung, and S. Park are with the Department of Electrical and Computer Engineering, Korea University, Seoul 02841, Republic of Korea (e-mails: \{kingdom0545,rupang1234,soohyun828\}@korea.ac.kr).}
    \thanks{S. Park is the corresponding author of this paper.}  
}
\maketitle

\begin{abstract}
The advent of reusable rockets has heralded a new era in space exploration, reducing the costs of launching satellites by a significant factor. Traditional rockets were disposable, but the design of reusable rockets for repeated use has revolutionized the financial dynamics of space missions. The most critical phase of reusable rockets is the landing stage, which involves managing the tremendous speed and attitude for safe recovery. The complexity of this task presents new challenges for control systems, specifically in terms of precision and adaptability. Classical control systems like the proportional-integral-derivative (PID) controller lack the flexibility to adapt to dynamic system changes, making them costly and time-consuming to redesign of controller. This paper explores the integration of quantum reinforcement learning into the control systems of reusable rockets as a promising alternative. Unlike classical reinforcement learning, quantum reinforcement learning uses quantum bits that can exist in superposition, allowing for more efficient information encoding and reducing the number of parameters required. This leads to increased computational efficiency, reduced memory requirements, and more stable and predictable performance. Due to the nature of reusable rockets, which must be light, heavy computers cannot fit into them. In the reusable rocket scenario, quantum reinforcement learning, which has reduced memory requirements due to fewer parameters, is a good solution.
\end{abstract}


\IEEEpeerreviewmaketitle

\section{Introduction}
The advent of reusable rockets, epitomized by SpaceX's Falcon 9, has revolutionized space exploration by significantly reducing the costs associated with launching rockets~\cite{spaceX}. Traditionally, rockets have been single-use items: after delivering their payloads, they either burn up in the Earth's atmosphere or fall into the sea. Reusable rockets are designed not to be used once and discarded, but to carry a payload (satellite) into space and then return to Earth for re-flying~\cite{SpaceX_2}. By reusing rockets, a significant reduction in the financial constraints has been observed, proving to be an indispensable asset to low Earth orbit (LEO) satellite communication systems that require a multitude of satellites for large coverage~\cite{tvt202205kim}. In fact, the use of reusable rockets has reduced the cost of launching LEO satellites by a factor of 20~\cite{cost_reduction}.
The on-flight control of reusable rockets during their own missions can be majorly divided into three stages, i.e., propulsion, on-flight control, and landing, where the three stages have been independently and actively discussed. However, the most important of the three stages is landing. Reusable rockets re-enter the Earth's atmosphere at tremendous speeds, often several times the speed of sound. In order to use the rocket again, it is necessary to control such tremendous speed and attitude, and controlling this is the landing technique. The landing stage is what enables this reusability by safely recovering the vehicle.

Reusable rockets, however, present new challenges for \textit{control systems}. Landing a rocket vertically is an incredibly complex task that requires precise control, timing, and coordination. This complexity can make it the most challenging part of the flight. The key characteristic of a reusable rocket is its ability to make a controlled landing after its mission. However, the safe landing of these reusable rockets is a complex operation. The rockets' speed, angular velocity, and the external influences, such as wind, need to be meticulously accounted for during their return to Earth. Control methods, to date, have predominantly relied on classical controls such as proportional-integral-derivative (PID) controller. The challenge with these methods is their lack of flexibility. 
 
 Classical control systems, despite their prevalent usage, are constrained by their lack of adaptability to dynamic changes within the system. Any alteration in the dynamics necessitates a comprehensive redesign of the controller, a process that can be time-consuming and costly. These challenges highlight the need for a more adaptable, efficient, and robust control methodology. Recent advancements in artificial intelligence and, specifically, reinforcement learning (RL) offer promising alternatives. RL introduces the unique advantage of adaptability, requiring only hyperparameter re-tuning rather than a full redesign of controller when system dynamics change. This is particularly useful for systems with complex dynamics that are difficult to accurately model, such as the landing of a reusable rocket, a system with non-linearity or uncertainty. This adaptability of RL to dynamic system changes suggests its potential applicability to the control systems of reusable rockets. However, integrating RL into a reusable rocket control system still requires many parameters for training. The need for many parameters for training means that there is still a risk of overfitting and the training process is very complex. In this challenge, \textit{quantum-based algorithms} offer good solutions. \textit{Quantum RL (QRL)}, application of quantum mechanics concepts to classical RL, offers potential benefits in terms of reducing the number of parameters required for learning, promising a more efficient approach.

 In classical RL, information is expressed in bits, but in QRL, the unit of information expression is a quantum bit (qubit). A qubit, unlike a classical bit, can exist in a superposition of states. This means it can represent multiple states simultaneously, potentially encoding more information than a classical bit. Consequently, this can reduce the number of parameters required to represent complex systems or problems, thus simplifying the learning process. Fewer parameters increase computational efficiency by reducing computational load and memory requirements. Also, since there are fewer dimensions of the parameter space to explore, fewer parameters allow for faster convergence during the learning process. In addition, fewer parameters can improve the robustness and stability of the learning process. In RL, it's often the case that small changes in parameters can lead to large changes in learned behavior. By reducing the number of parameters, the system can become less sensitive to parameter variations, leading to more stable and predictable performance.

\BfPara{Contributions}
The main contributions are as follows.

\begin{itemize}
    \item \textbf{First application of QRL to reusable rocket landing and reducing the number of parameters.} This study pioneers the application of QRL to reusable rocket control systems. By minimizing the parameters necessary for learning, QRL could dramatically decrease the computational requirements usually associated with classical RL. 
    \item \textbf{High application potential to other flight control systems.} This work aims to introduce the advantages of QRL to the realm of reusable rockets. This study suggests high applicability to complex and non-linear control systems. Because the same algorithms are needed when a space probe lands on the moon or Mars, QRL will be of great help in humanity's challenge to space.
\end{itemize}

\section{Related Work}
As a novel control method for precise landing, one study enables the training of a neural network controller for rocket engines using RL \cite{rocket_2}. 
By employing RL, the controller can interact with suitable simulation environments and infer optimal flight paths for various scenarios to achieve accurate landings.
In another study, a deep RL (DRL) approach is proposed to optimize the continuous ignition phase of a general gas generator engine \cite{rocket_3}.
This approach, unlike carefully tuned open-loop sequences and PID controllers, exhibits superior performance as the learned policies can reach different steady-state operating points and adapt persuasively to changing system parameters.
On the other hand, both papers only consider the rocket's engine in their research.
Furthermore, there have been cases where RL was applied to traditional control algorithms to create more sophisticated algorithms.
In \cite{rocket_1}, the objective is to improve PID and the model predictive control using deep deterministic policy gradient (DDPG).
While the classical controller achieves acceptable performance, DDPG showcase convergence towards a more stable and consistent rocket landing.
Furthermore, there have been cases where RL was applied to the guidance algorithm.
In \cite{rocket_4}, a real-time feedback-capable guidance algorithm was developed by applying DRL to the guidance algorithm to ensure fuel optimization and convergence.
This paper presents a model pre-training framework based on imitation learning, aiming to enhance model convergence by fitting it with optimal data.
The above papers aimed to address the stability of rocket landing using existing RL methods.
However, in the case of artificial neural networks, once the process is completed with the given hyperparameters, they cannot be changed, limiting their adaptability for further expansion.
Especially when the state-action space becomes large, traditional RL methods like Q-learning become impractical to use.

\section{Quantum Neural Network}

Qubits are the fundamental units of information in quantum systems and can be represented using the base states $\left| 0 \right>$ and $\left| 1 \right>$. They can also be expressed as superpositions of 0 and 1 \cite{qubit_1}.
A single qubit state can be expressed as a normalized 2D complex vector as, 
\ $\left| \psi  \right>=\alpha \left| 0  \right>+\beta \left| 1  \right>,\left\| \alpha \right\|^{2}+\left\| \beta \right\|^{2}=1$, 
\label{eq:qubits_1}
where $\left\| \alpha \right\|^{2}$ and $\left\| \beta \right\|^{2}$ represent the probabilities of observing $\left| 0 \right>$ and $\left| 1 \right>$, respectively, from the qubit \cite{iotj23park}.
To represent this in the Bloch sphere, it can be geometrically expressed as,
$\left| \psi  \right>=\cos(\frac{\theta }{2})\left| 0  \right>+e^{i\phi }\sin(\frac{\theta }{2})\beta \left| 1  \right>(0\leq \theta , \phi \leq \pi )$, 
where $\theta$ is a parameter that determines the probabilities of measuring $\left| 0 \right>$ and $\left| 1 \right>$, and $\phi$ represents the relative phase.
Furthermore, qubits can become entangled with each other, resulting in a strong correlation between the two separate qubits.
Through these features, using qubits allows for the control of more information than what is possible with classical bits.
When assuming a $q$ qubit system, the quantum states existing in the system's Hilbert space are represented as $ \left| \psi \right>=\omega _{1}\left| 0\cdots 0 \right>+\cdots+\omega _{2^{q}}\left| 1\cdots 1 \right>$,
where $\omega $ is a complex number representing the probability amplitude for each base, and $\psi$ denotes the quantum state.
To design and train QNN effectively, qubits must be controllable.
Control over the position of these qubits is achieved by using basic quantum gates.
One of the representative examples of basic quantum gates is the rotation gate, expressed as $R_x$, $R_y$, and $R_z$. These gates perform rotations around the $x$-, $y$-, and $z$- axes, respectively.
These gates not only control qubits but also encode classical data into quantum states.
The basic quantum rotation gates are single-qubit gates that can only be applied to individual qubits. However, there are also multi-qubit gates that act on two or more qubits simultaneously.
Based on the above theory and concepts, various types of gates are assembled to construct a QNN model, and quantum reinforcement learning (QRL) is carried out using this model as a basis.
Additional explanations for the three components of QNN are provided in the following paragraph.

\begin{figure}
  \centering
  \includegraphics[width=0.9\linewidth]{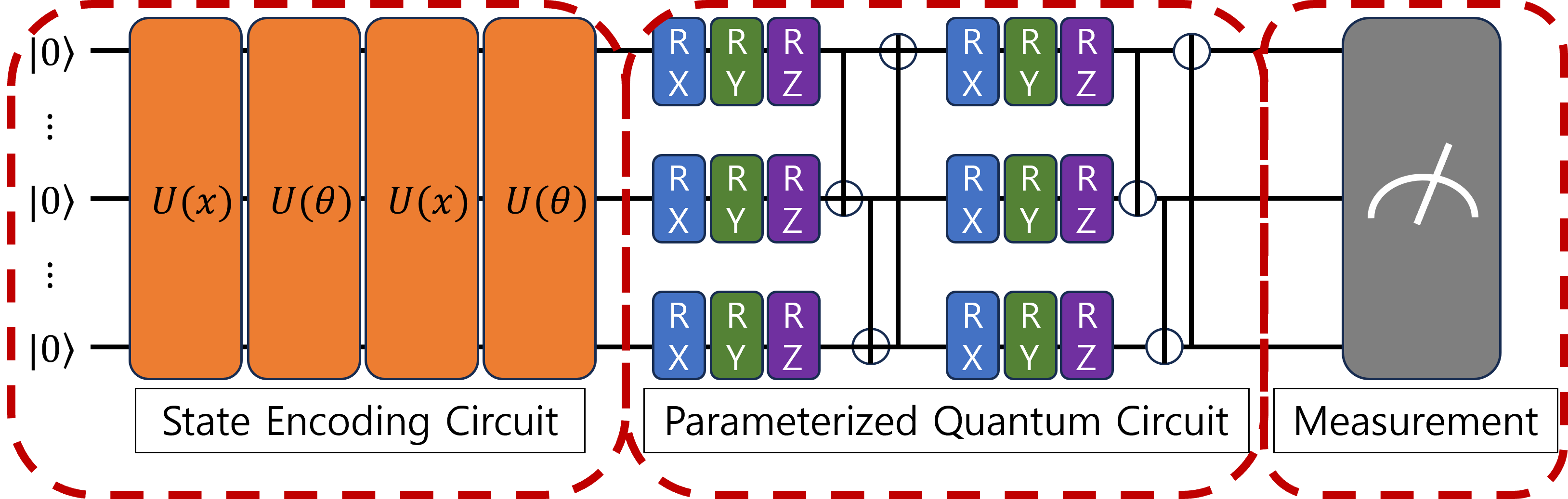}
  \caption{Architecture of quantum neural network (QNN).}
  \label{fig:Quantum Reinforcement Learning}
\end{figure}

\BfPara{State Encoding}
The function of the encoder is to encode classical data into quantum states. Quantum circuits cannot directly take classical bits as input, which is why an encoder is used to perform this conversion.
As a result, the state encoder utilizes classical data represented by parameters to transform $q$ number of $\left| 0 \right>$ states into qubits. 
This transformation is achieved by passing the states through an array of rotation gates and entanglement circuits.

\BfPara{Parameterized Quantum Circuit (PQC)}
PQC performs calculations similar to the multiplication in accumulated hidden layers of classical neural networks.
It accomplishes the desired computations.
In the PQC layer, the quantum state output from the encoder is used as input. Quantum gates are then utilized to rotate the state at specific angles, providing the required values for actions and state-related information.
The structure of QNN is shown in Fig.~\ref{fig:Quantum Reinforcement Learning}.

\BfPara{Measurement}
The quantum state obtained through PQC becomes the input for the measurement layer.
In this layer, quantum data are decoded back to the original data for optimization purposes, and then the measurement is performed on the input.
The measurement operation is akin to performing matrix multiplication with the projection matrix along the $z$- axis.
The most commonly used axis in measurements is the z-axis, but it can also be any other appropriately defined direction.
After measuring the quantum state, the quantum state collapses and its properties become \textit{observable}.
After carrying out the decoding process, \textit{observable} is utilized to minimize the loss function.
After that, it is necessary to differentiate for backpropagation.
By the way, applying the chain rule causes the state of the qubit to collapse completely, making it impossible to distinguish quantum data.
Therefore, the technique of obtaining the loss gradient in QNN training is through the symmetric difference quotient of the loss function with respect to observable.

\section{Algorithm Design}\label{sec:Algorithm Design}

\subsection{Reusable Rocket Landing}

\begin{figure}
  \centering
  \includegraphics[width=1\linewidth]{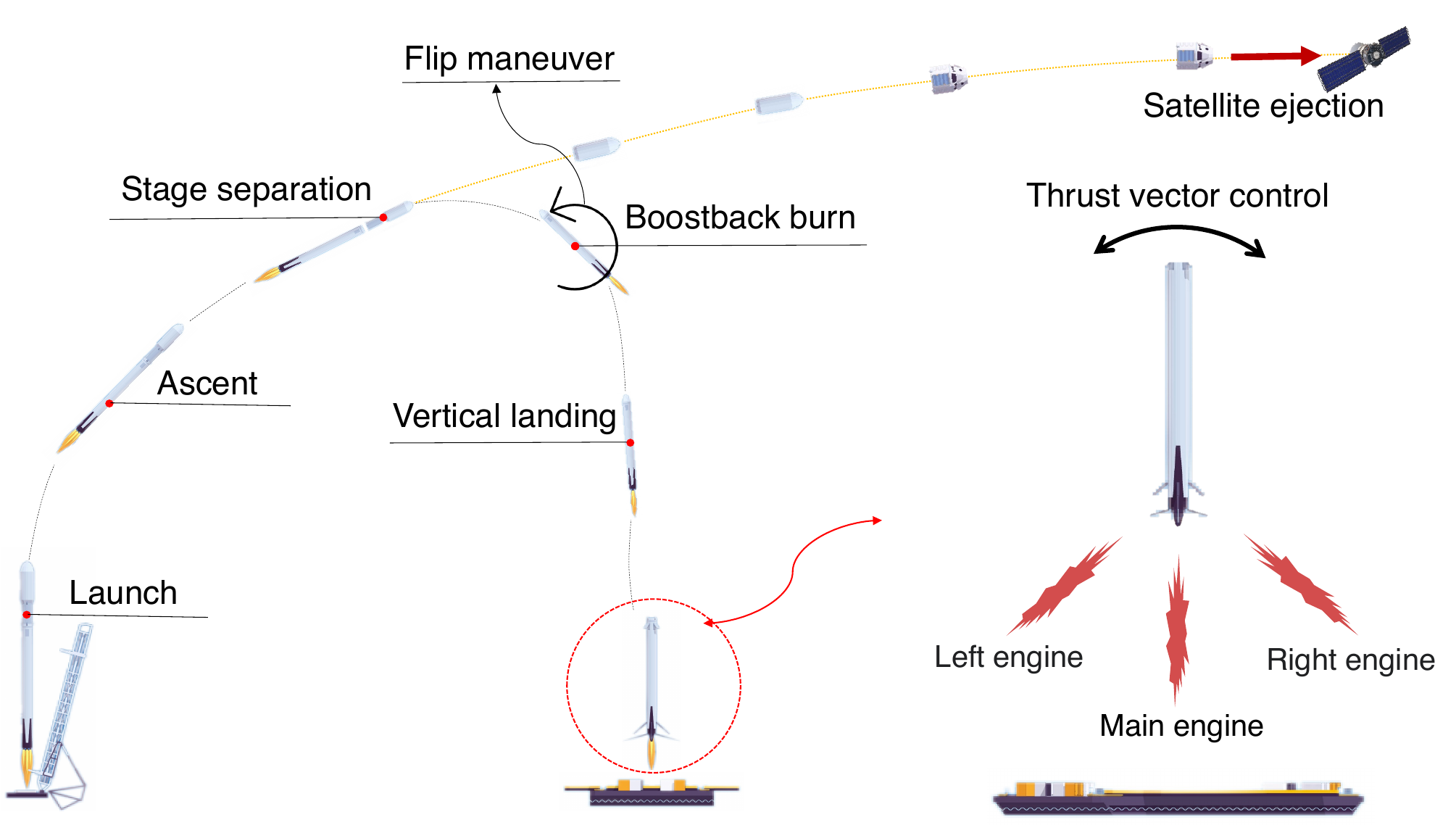}
  \caption{Depiction of reusable rocket landing trajectory.}
  \label{fig:Depiction of reusable rocket landing trajectory}
\end{figure}

This chapter describes the process of launching a rocket and returning it to Earth for reuse. The landing procedure of the reusable rocket is explained in the Fig.~\ref{fig:Depiction of reusable rocket landing trajectory}.

\begin{enumerate}
    \item \textbf{Launch.} This begins with the ignition of the rocket's engines, creating thrust to overcome Earth's gravity. The rocket lifts off from the launch pad, and the engines continue to burn fuel to propel the rocket upwards.
    \item  \textbf{Ascent.} During the ascent phase, the rocket follows a predetermined trajectory to reach the desired orbit. The engines maintain a specific thrust profile to ensure optimal fuel efficiency and alignment with the target orbit. Aerodynamic forces and engine gimballing are carefully managed to steer the rocket.
    \item \textbf{Stage Separation.} Most rockets are designed with multiple stages, each containing its own engines and fuel. Stage separation occurs when the lower stage has expended its fuel, and it is jettisoned to reduce weight. This allows the upper stage(s) to ignite and continue the journey with a fresh supply of fuel. Stage separation must be precisely timed and controlled to prevent collision or misalignment.
    \item \textbf{Flip Maneuver.} In reusable rocket systems, the flip maneuver is often performed by the first stage after separation to prepare for reentry and landing. This involves rotating the stage to position the engines downward, readying it for the controlled descent back to Earth.
    \item \textbf{Boostback Burn.} The boostback burn is executed to adjust the trajectory of the returning stage so that it lands at the designated recovery area. It may involve a series of controlled engine burns to slow down and guide the stage towards its landing target.
    \item \textbf{Satellite Ejection.} Once the rocket reaches the designated orbit, it deploys the satellite or payload. This requires careful alignment and timing to ensure the satellite is placed into the correct orbit. This must be gentle enough to avoid damaging the payload but forceful enough to ensure proper deployment.
    \item \textbf{Vertical Landing.} This final phase is the controlled descent and vertical landing. This requires a series of carefully coordinated engine burns to slow down the rocket as it approaches the landing site. The rocket's landing legs are deployed, and the engines are used to perform a controlled touchdown. 
In this case, robust control technology is required. The reusable rocket controls the thrust vector through the side engine and the main engine to safely land on the landing pad.
\end{enumerate}

\subsection{Motivation for Quantum-based Approach}
In modern DRL applications, quantum-based approaches have been considered based on its supremacy. In this paper, a QRL algorithm is considered due to efficient parameter utilization and its corresponding discussions are as follows.
\begin{itemize}
    \item Firstly, quantum entanglement is a phenomenon where two or more qubits become linked and the state of one qubit can instantaneously affect the state of another, no matter how far apart they are. This correlation can be used to encode complex relationships with fewer parameters than would be needed in classical systems. 
    \item Furthermore, quantum parallelism allows quantum computers to evaluate a large number of possibilities at once. This capability might reduce the need for a large parameter space, as quantum systems can examine multiple solutions concurrently.
    \item Lastly, quantum systems can compactly represent and manipulate high-dimensional vectors and matrices. This capability could allow for a more efficient representation of the parameter space.
\end{itemize}

\subsection{QRL-based Rocket Landing Control}

QRL is an emerging field that combines principles from quantum mechanics and classical RL to address complex decision-making problems in quantum systems. QRL, much like traditional RL, revolves around three key components: state, action, and reward.

In QRL, states represent the observable quantum system that an agent interacts with.
In contrast to classical RL, where states are represented as classical variables, QRL describes states using quantum states that exist in a high-dimensional vector space known as Hilbert space.
These quantum states are represented by unit vectors, and they form a complete orthonormal basis, allowing for a full description of the quantum system.
We refer to these quantum states as eigen states.
In this case, the selection of observable quantities is crucial because they directly influence the actions an agent will encounter and the rewards it will receive.
Additionally, actions are also referred to as eigen actions, representing the transformations that the agent can apply to the quantum system.
These actions are represented by quantum gates, which are unitary operators that act on the quantum states. 
Quantum gates allow the agent to manipulate the quantum system's state, enabling it to navigate through the Hilbert space to perform various quantum operations.
Quantum actions differ from classical actions, as they can leverage quantum phenomena like superposition and entanglement, providing unique opportunities for exploration and exploitation in decision-making processes.
The reward function in QRL is similar in concept to traditional RL, as it represents the feedback an agent receives from the environment after performing an action in a specific state.
The agent aims to maximize the cumulative reward over time, driving it to learn effective policies for quantum tasks.
However, the implementation of the reward function in QRL might require specialized techniques due to quantum characteristics.

\subsubsection{State} \label{sec:state}

\begin{figure}
  \centering
  \includegraphics[width=0.748\linewidth]{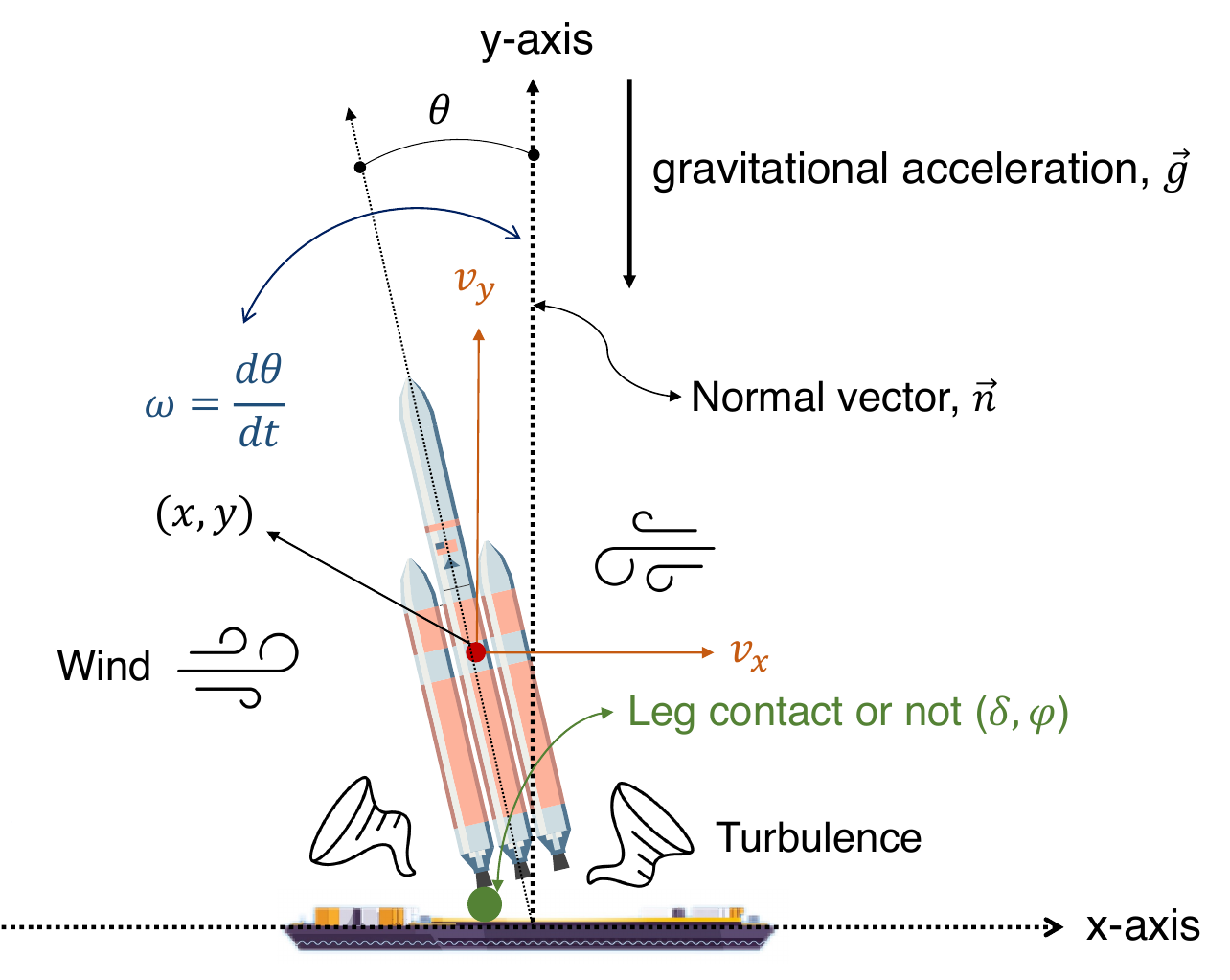}
  \caption{Description of states.}
  \label{fig:State}
\end{figure}

The experimental environment is a trajectory optimization problem for reusable rocket landing. Fig.~\ref{fig:State} describes the states in the experimental environment. The coordinates of the landing pad where the reused rocket should land after returning to Earth are always $(0,0)$. The observation space of the reuse rocket is expressed as an 8-dimensional vector, which is the current $x$-coordinate, $y$-coordinate of the reuse rocket, its linear velocities in $x$ \& $y$, its angle, its angular velocity and two booleans indicating whether both legs of the reuse rocket are in contact with the ground. As a result, the state is expressed as an 8-dimensional vector as,
$\mathcal{S}\triangleq [x, y, v_x, v_y, \theta, \omega, \delta , \varphi ]$, 
where $x$ and $y$ represent the location coordinates of the current reusable rocket when the coordinates of the landing pad are $(0,0)$. The $v_x$, $v_y$ represent the current velocity of the reusable rocket for each axis. The $\theta$ means the angle formed by the normal vector ($\overrightarrow{n}$) of the ground and the reusable rocket. The $\omega$ is the angular velocity of the reusable rocket, indicating how much the rocket's angle changes over time. The $\delta$ and $\varphi$ are booleans representing whether both legs of the reuse rocket are in contact with the ground.
 Additionally, environmental factors include wind, turbulence, and gravitational acceleration. 
In the experimental environment, the reusable rocket lands in a windy environment. The wind strength is randomly applied below the maximum wind strength.
 And disturbance represents a rotational wind that can affect the lander's orientation and control. rotational wind refers to a turbulent effect applied to the lander during its descent. It's like a turbulent wind that applies a rotational force to the reusable rocket, making it more challenging to maintain a stable descent and landing. The effect would simulate what might happen if the reusable rocket were caught in a gust of wind that caused it to rotate.
 Earth's gravitational acceleration is applied to the acceleration.

\subsubsection{Action}

A total of four actions can be taken when a reused rocket lands after returning to Earth, which is expressed as, 
$\mathcal{A}\triangleq\{\textit{Do-Nothing}, \textit{Left-Engine}, \textit{Right-Engine}, \textit{Main-Engine}\}$. Fig.~\ref{fig:Depiction of reusable rocket landing trajectory} shows the actions that rockets can take. The reusable rocket observes its own state and takes one of the four actions from $\mathcal{A}$. Firstly, the reusable rocket continues with its current motion, with no additional thrust. Then, applies thrust to the right and left to control the horizontal position. Lastly, applies thrust upward to slow the descent.

\subsubsection{Reward}\label{sec:reward}
A reusable rocket takes a specific action in a specific state and receives a reward according to the reward function. The reusable rockets should maintain a good attitude in the air until just before landing and reach the landing pad as quickly as possible. In the landing scenario of a reuse rocket, the reward is defined as,
$\Re = -\xi \times (d_t-d_{t-1})-\mu \times (v_t-v_{t-1})-\chi \times(\omega _t-\omega _{t-1}) + \kappa$ 
where $d_t$, $d_{t-1}$, $v_t$, $v_{t-1}$, $\omega_t$, and $\omega_{t-1}$ are the distance to the landing pad at time $t$ and $t-1$, the velocity of the reusable rocket at time $t$ and $t-1$, the angular velocity at time $t$ and $t-1$, respectively. $\xi$, $\mu$, $\chi$ are correction constants for distance, velocity and angular velocity, respectively. $\kappa$ is a boolean value indicating whether the reused rocket landed smoothly on the landing pad. Also, $d_t$ and $v_t$ are expressed as $d_t=\sqrt{x_t^2+y_t^2}$ and $v_t=\sqrt{v_x^2+v_y^2}$.
This reward induces the rocket to slow down to reduce its distance to the landing pad and land softly. It keeps the angular velocity to a minimum to prevent rolling moments and induces it not to take off again after landing.

\section{Performance Evaluation}

\begin{table}[t!]
\scriptsize
\centering
\caption{Environmental hyperparameters}
\renewcommand{\arraystretch}{1.0}
\begin{tabular}{l||r}
\toprule[1pt]
\textbf{Notation} & \textbf{Value} \\ \midrule
Dense layer, $L$ & 3 \\
Discount factor, $\gamma$ & 0.99 \\
The number of nodes in all layers & 64 \\
Learning rate & 0.0005 \\
Training epochs & 10k \\
Activation function and optimizer & ReLU and Adam\\
\bottomrule[1pt]
\end{tabular}
\label{tab:hyperparameter}
\end{table}

The experimental environment is \textit{Lunar-Lander} provided by Open AI gym. 
In this paper, a classical RL-based neural network consisting of $L$ dense layers is considered. Also, due to the decaying epsilon greedy method, reusable rockets can experience different behaviors in different states. Additionally, the number of qubits used for training, measurement, and gate are 4, Pauli-$Z$, and $X$-rotations, respectively.
Table~\ref{tab:hyperparameter} summarizes environmental hyperparameters in simulations.

\begin{figure}[t!]
    \centering
    \includegraphics[width=0.9\linewidth]{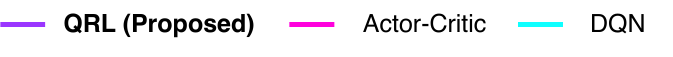}\\
    \includegraphics[width=0.70\linewidth]{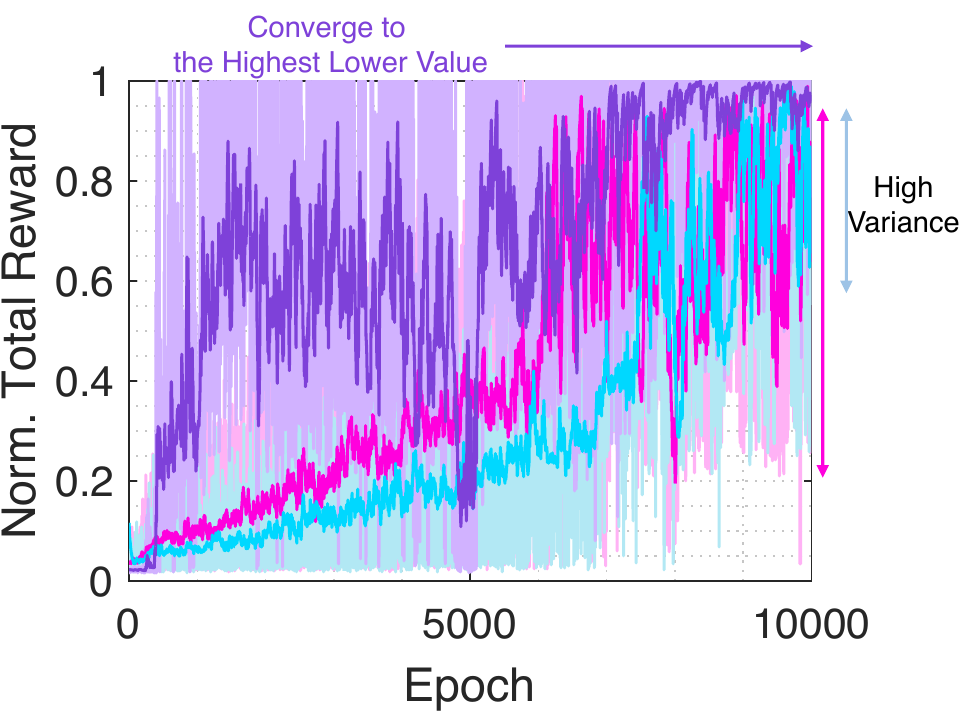}
    \caption{Reward of the proposed and all benchmark algorithms over unit time.}
    \label{fig:Reward}
\end{figure}

\begin{figure}[t!]
    \centering
    \includegraphics[width=0.9\linewidth]{Figure4_label.pdf}\\
    \subfigure[Total Loss.]{
    \includegraphics[width=0.45\linewidth]{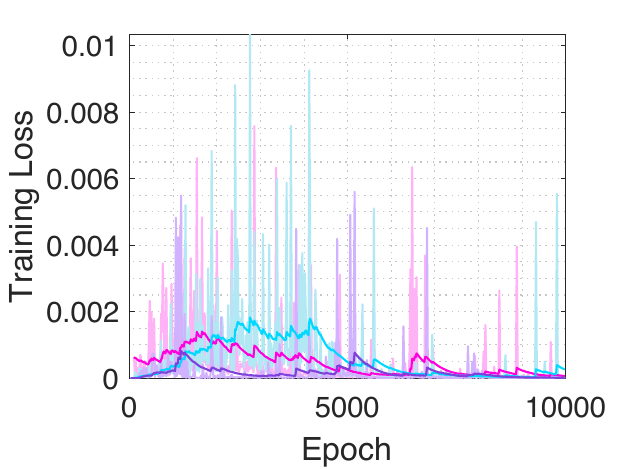}
    }
    \subfigure[Enlarged Loss.]{
    \includegraphics[width=0.45\linewidth]{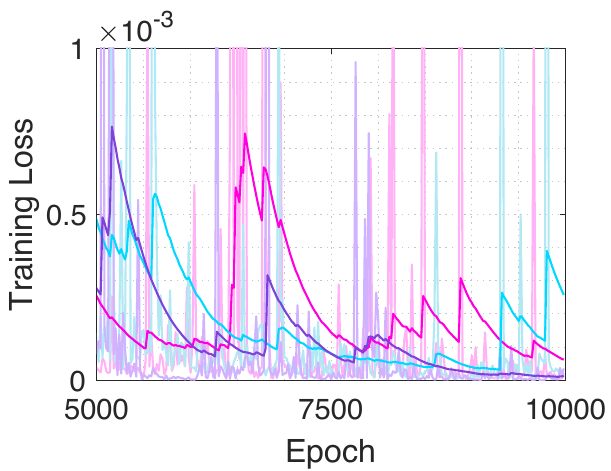}
    }
    \caption{Training Loss.}
    \label{fig:loss}
\end{figure}

\BfPara{Rewards and Training Losses}
Firstly, training performance is evaluated in terms of reward value and convergence. Fig.~\ref{fig:Reward} shows the reward per epoch of the proposed algorithm and benchmarks. The purple-line, pink-line, and blue-line are for QRL (proposed), \textit{actor-critic}, and \textit{deep Q-network (DQN)}. In the reusable rocket landing scenario, it can be seen that the proposed algorithm increases the reward the fastest and obtains the highest convergence reward. And the convergence of the reward also appears the fastest. 
Variation is also small for the proposed algorithm compared to other benchmarks.
Therefore, QRL outperforms conventional classical RL algorithms, proving that QRL is an effective solution for safe landing of reusable rockets.
Secondly, the training loss over epoch is presented in Fig~\ref{fig:loss}. Fig.~\ref{fig:loss}(a) shows the loss trend of the proposed algorithm and other benchmarks during the whole epoch, and Fig.~\ref{fig:loss}(b) shows the enlarged loss from 5000 to 10000 epoch. The proposed algorithm and other benchmarks all decrease loss as the epoch progresses. However, Fig.~\ref{fig:loss}(b) shows the proposed algorithm eventually converges to the lowest training loss. The loss of other benchmarks converges to larger values and fluctuates.

\BfPara{Number of Parameters}
Table~\ref{tab:The number of Normalized Parameters according to the Algorithm} shows the normalized number of parameters of the proposed algorithm and other benchmarks. It can be seen that the proposed algorithm has the smallest number of parameters for training. 
The normalized ratios of numbers of parameters in the proposed algorithm are $4.29$ and $3.77$ times smaller than \textit{DQN} and \textit{actor-critic}.
This is suitable for landing control of reusable rockets, where efficient use of parameters is important due to the limited computation.

\begin{table}[!t]
\scriptsize
    \centering
    \caption{The number of Normalized Parameters}
    \begin{tabular}{l||c|c|c}
    \toprule
     & \textbf{QRL (Proposed)} & {DQN} & {Actor-Critic}\\
    \midrule
    \# Normalized Parameters & $\mathbf{0.233}$ & 1 & 0.879\\
    \bottomrule
    \end{tabular}
    \label{tab:The number of Normalized Parameters according to the Algorithm}
\end{table}

\section{Conclusions}\label{sec:conclusion}
As proposed in this paper, QRL-based adaptive control has shown promise in the stabilized landing of reusable rockets, by reducing parameters and improving efficiency. 
This research for the demonstration QRL-based stabilized rocket landing's efficacy, aligning with rocket's computation/memory constraints.
Based on performance evaluation, it can be verified that our proposed QRL-based algorithm achieves desired performance improvements.
\input{citation.bbl}

\end{document}